\documentclass[conference]{inc/IEEEtran}
\usepackage{times}

\usepackage[numbers]{natbib}
\usepackage{multicol}
\usepackage[bookmarks=true]{hyperref}

\usepackage{amsmath}
\usepackage{amssymb}
\usepackage{booktabs}
\usepackage{graphicx}
\usepackage{afterpage}  
\usepackage{dblfloatfix} 
\usepackage[section]{placeins} 
\usepackage{placeins}
\usepackage{float}
\usepackage{cuted}
\usepackage{caption}
\usepackage{algorithm}
\usepackage{algpseudocode}
\usepackage{xcolor}

\DeclareMathOperator*{\argmin}{arg\,min}

\usepackage{comment}
\usepackage{lipsum}

\usepackage{pifont}
\usepackage{xcolor}
\newcommand{\cmark}{{\color{green!70!black}\ding{51}}}
\newcommand{\xmark}{{\color{red}\ding{55}}}

\hypersetup{
    colorlinks=true,
    urlcolor=magenta
}

\begin{document}

\title{TinySDP: Real Time Semidefinite Optimization\\for Certifiable and Agile Edge Robotics}

\author{
\authorblockN{
Ishaan Mahajan$^{1}$,
Jon Arrizabalaga$^{2,\ddagger}$,
Andrea Grillo$^{3, 4, \ddagger}$,
Fausto Vega$^{2,\ddagger}$,\\
James Anderson$^{1,\dagger}$,
Zachary Manchester$^{2,\dagger}$,
and Brian Plancher$^{3}$
}

\authorblockA{
$^{1}$Columbia University, USA \quad
$^{2}$Massachusetts Institute of Technology, USA \quad
$^{3}$Dartmouth College, USA \quad \\
$^{4}$École Polytechnique Fédérale de Lausanne (EPFL), Switzerland \quad \\
$^{\ddagger}$Equal contribution \quad
$^{\dagger}$Equal advising
}
}

\maketitle

\begin{abstract}
    Semidefinite programming (SDP) provides a principled framework for convex relaxations of nonconvex geometric constraints in motion planning, yet existing solvers are too computationally expensive for real-time control, particularly on resource-constrained embedded systems. To address this gap, we introduce TinySDP, the first semidefinite programming solver designed for embedded systems, enabling real-time model-predictive control (MPC) on microcontrollers for problems with nonconvex obstacle constraints. Our approach integrates positive-semidefinite cone projections into a cached-Riccati-based ADMM solver, leveraging computational structure for embedded tractability. We pair this solver with an \emph{a posteriori} rank-1 certificate that converts relaxed solutions into explicit geometric guarantees at each timestep. On challenging benchmarks, e.g., cul-de-sac and dynamic obstacle avoidance scenarios that induce failures in local methods, TinySDP achieves collision-free navigation with up to 73\% shorter paths than state-of-the-art baselines. We validate our approach on a Crazyflie quadrotor, demonstrating that semidefinite constraints can be enforced at real-time rates for agile embedded robotics. Project Website: \url{https://a2r-lab.org/TinySDP/}
\end{abstract}

\IEEEpeerreviewmaketitle

\section{Introduction}


\begin{figure*}[t]
    \centering
    \includegraphics[width=\textwidth]{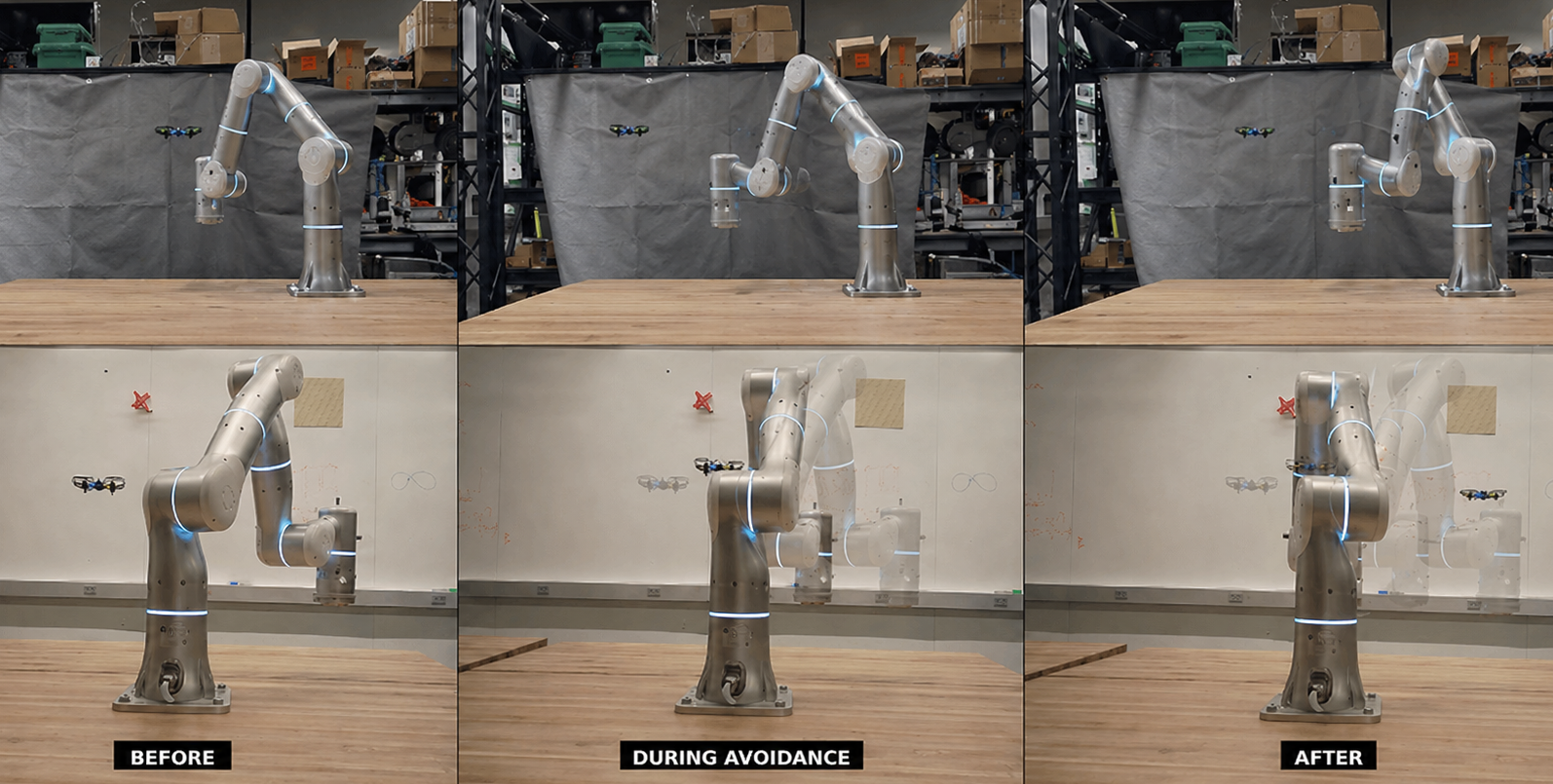}
    \caption{
    Front (top) and side (bottom) views of the Crazyflie running TinySDP online to avoid a moving arm. Left to right: before, during, and after avoiding the moving obstacle.
    }
    \label{fig:hardware}
    \vspace{-15pt}
\end{figure*}


Safe motion planning and control in dynamic environments remains a challenge in robotics. Model-predictive control (MPC) is attractive for this setting due to its ability to reason about system dynamics and constraints (e.g., obstacle avoidance), and has been demonstrated on a wide range of robotic problems across numerous tasks and scenarios \cite{kalman1960lqr,darby2012mpc,falanga2018pampc,torrente2021data,mcguire2019minimal,narkhede2022sequential,tinympc,chiu2022collision,ammour2022mpc,kuindersma2016optimization,liu2016entry,altroc}.

Recent advances in embedded optimization have expanded the scope of MPC on resource-constrained platforms~\cite{ferreau2017embedded, kouzoupis2015first, jerez2014embedded}, including the development of popular packages, e.g., OSQP~\cite{stellato_osqp_2020}, CVXGEN~\cite{mattingley2012cvxgen}, ECOS~\cite{ecos}, SCS~\cite{scs}. In particular, TinyMPC~\cite{tinympc} demonstrated that constrained quadratic MPC can be solved at high rates on microcontrollers by exploiting the Riccati structure through a combination of offline caching and first-order methods. Building on this foundation, subsequent work~\cite{conic-tinympc,Mahajan2025adaptiveTinyMPC} improved the robustness of the solver and added support for second-order cone constraints. 
Despite this progress, obstacle avoidance with formal guarantees remains an open problem in embedded MPC, particularly because obstacle constraints are inherently nonconvex. As such, existing real-time MPC implementations rely on conservative over-approximations or local convexifications that fail to capture global geometric structure and are fragile in practice (e.g., linearized distance constraints and tangent half-spaces~\cite{richards2002aircraft, schouwenaars2001mixed, lozano1983spatial, tinympc}). This is because solutions that satisfy such approximations can still tunnel through obstacles, requiring extensive heuristic tuning of scenario-specific safety margins, and often still fail in the presence of narrow passages, cul-de-sacs, or dynamic obstacles~\cite{zeng2021safety, thirugnanam2022safety}. 

Similarly, while barrier function based methods are also widely used for obstacle avoidance and collision avoidance, crafting effective barrier functions also relies on significant tuning~\cite{ames2016control, xiao2021high, wieland2007constructive}. And, while Hamilton Jacobi (HJ) reachability tools can be used to construct tuning-free barrier functions, these tools do not scale well to high-dimensional systems~\cite{tonkens2022refining, choi2021robust}.
More recently, learning-based barrier functions leveraging neural networks have been developed to overcome the dual scalability and tuning challenges. However, deploying the resulting moderate-sized neural networks for real-time applications is often infeasible on embedded platforms~\cite{pmlr-v164-dawson22a, liu2023safe}.

In contrast, semidefinite programming (SDP) provides a principled way to pose nonconvex geometric constraints as convex by lifting quadratic or polynomial constraints into higher-dimensional positive semidefinite cones, and has been successfully applied in offline motion planning and trajectory optimization~\cite{kang2025global, parrilo2005exploiting, henrion2005positive, majumdar2017funnel, marcucci2023motion, dong2026fast, papalia2025certifiable}, globally optimal state estimation and localization~\cite{dumbgen2024toward, groudiev2025sampling}, and sum-of-squares optimization with tools like SOSTOOLS~\cite{papachristodoulou2013sostools}. However, as with many of the prior techniques, general-purpose SDP solvers are computationally expensive and memory-intensive. As such, they are generally poorly suited for real-time receding horizon control, particularly for embedded applications~\cite{vandenberghe1996semidefinite, marcucci2023motion}. 

In this work, we show that it is possible to overcome these challenges through careful exploitation of problem structure. 
We introduce TinySDP, the first semidefinite programming solver designed for real-time obstacle avoidance on embedded systems. Our approach uses a per-stage, convex, semidefinite relaxation of nonconvex, quadratic, disk-based obstacle avoidance constraints. 
These relaxations are small, structured, and compatible with Riccati-based MPC solvers, including cached variants~\cite{tinympc}.
We pair this with a simple rank-1 certificate that converts the relaxed solution into an explicit geometric safety guarantee at every timestep. When the certificate holds, it provably certifies that the true robot position avoids all obstacles, despite the use of an efficient convex relaxation. 

Our approach builds on prior observations that lifted convex relaxations are empirically tight in structured control problems~\cite{bamieh2024linear}. In particular, lifted semidefinite formulations of linear-quadratic control problems recover low-rank solutions, providing exact certificates of optimality~\cite{boumal2020deterministic, ma2018implicit}. 

We evaluate TinySDP on two challenging benchmark tasks that are designed to expose common failure modes for standard formulations: 1) a static U-shaped cul-de-sac and, 2) multiple dynamic moving-gap scenarios. Through systematic comparisons, we show that our method is less conservative, with path lengths up to 73\% shorter than baseline approaches, while also remaining certifiably collision-free at each timestep through the use of our a posteriori rank-1 safety certificate. 
We also implement the full solver within the TinyMPC framework~\cite{tinympc}, and validate our algorithm through real-world hardware deployments on a Crazyflie 2.1 Brushless quadrotor at 25~Hz control rates, demonstrating that semidefinite constraints can be enforced at real-time rates for embedded robotic tasks.

\section{Background} \label{sec:background}
We review prior work on embedded MPC, with emphasis on cached Riccati-based solvers and operator-splitting methods, as well as convex relaxations for nonconvex quadratic constraints, providing the foundation for TinySDP (Section~\ref{sec:method}).

\subsection{Linear, Cached, and Riccati-Based MPC Solvers}

We highlight the components of the generic finite-horizon MPC problem that prevent a closed-form solution.

Consider a discrete-time linear time-invariant system with state
$x_k \in \mathbb{R}^n$ and control input $u_k \in \mathbb{R}^m$ at timestep $k$,
\begin{equation*}
x_{k+1} = A x_k + B u_k ,
\end{equation*}
where $A \in \mathbb{R}^{n \times n}$ and $B \in \mathbb{R}^{n \times m}$ define the system dynamics.

Given a horizon length $N$, and assuming quadratic stage and terminal costs, for $Q \succeq 0$, $R \succ 0$, and $Q_f \succeq 0$,  define
\begin{equation*}
\ell(x_k,u_k) = x_k^\top Q x_k + u_k^\top R u_k, \quad
\ell_f(x_N) = x_N^\top Q_f x_N,
\end{equation*}
and admissible convex state and control sets, $\mathcal{X}, \mathcal{U}$, the finite-horizon MPC problem is:
\begin{subequations}\label{eq:mpc}
\begin{align}
\min_{\{x_k,u_k\}} \quad &
\ell_f(x_N) + \sum_{k=0}^{N-1} \ell(x_k,u_k) \\[-0.25em]
\text{s.t.}\quad &
x_{k+1} = A x_k + B u_k, \\[-0.25em]
& {u_k \in \mathcal{U},} \quad {x_k \in \mathcal{X}}. \label{eq:mpc_ineq}
\end{align}
\end{subequations}

In the absence of the constraints ~\eqref{eq:mpc_ineq}, the problem reduces to an unconstrained linear-quadratic regulator (LQR), which admits a closed-form solution via the discrete-time Riccati recursion~\cite{lewis12optimal}, producing an affine control law: 
\begin{equation} \label{eq:lqr_forward}
    u_k = -K_k x_k - d_k,
\end{equation}

\begin{equation} \label{eq:lqr_backward}
    \begin{aligned}
    K_k &= (R + B^\top P_{k+1} B)^{-1} B^\top P_{k+1} A, \\[-0.25em]
    d_k &= (R + B^\top P_{k+1} B)^{-1} (B^\top p_{k+1} + r_k), \\[-0.25em]
    P_k &= Q + A^\top P_{k+1} A - A^\top P_{k+1} B K_k, \\[-0.25em]
    p_k &= q_k + A^\top p_{k+1} - A^\top P_{k+1} B d_k.
    \end{aligned}
\end{equation}

The presence of constraints destroys this structure, necessitating a numerical approach. To overcome this challenge, recent approaches have leveraged the alternating direction method of multipliers (ADMM)~\cite{boyd2011admm}, to separate the inequality constraints from the remainder of the problem, enabling the reintroduction of the efficient Riccati recursion. This is done by first introducing auxiliary variables, $z_k$, into the problem.

For clarity, and without loss of generality, we assume that the only inequality constraints in our MPC problem are on the controls, $u$. We can then introduce $\mathcal{I}_{\mathcal{U}}$, the indicator function of the admissible input set $\mathcal{U}$, and transform~\eqref{eq:mpc} into:
\begin{subequations}
    \begin{align}
    \min_{\{u_k,z_k\}} \quad &
    \ell_f(x_N) + \sum_{k=0}^{N-1} \ell(x_k,u_k)
    + \mathcal{I}_{\mathcal{U}}(z_k) \\[-0.25em]
    \text{s.t.}\quad &
     x_{k+1} = A x_k + B u_k, \\[-0.25em]
    & u_k - z_k = 0.
    \end{align}
\end{subequations}
If we then introduce $y_k$ as scaled dual variables and $\rho > 0$ as a penalty parameter, the augmented Lagrangian associated with the consensus constraint $u_k = z_k$ becomes:
\begin{equation*}
\underbrace{\ell_f(x_N) + \sum_{k=0}^{N-1}
\left(
\ell(x_k,u_k)
+ \mathcal{I}_{\mathcal{U}}(z_k)
+ \frac{\rho}{2}\|u_k - z_k + y_k\|_2^2
\right).}_{\mathcal L_\rho}
\end{equation*}
We then perform alternating minimization on $\mathcal L_\rho$ with respect to $x,u,z$, arriving at the three-step ADMM~\cite{boyd2011admm} iteration,
\begin{subequations}
\begin{align}
    \textbf{primal update}: x^{+},u^{+} &= \argmin_{x,u} \mathcal{L}_\rho(x,u,z,y) \label{eq:primal_update} \\[-0.5em]
    \textbf{slack update}: z^{+} &= \argmin_z \mathcal{L}_\rho(x^{+},u^{+},z,y) \label{eq:slack_update} \\[-0.5em]
    \textbf{dual update}: y^{+} &= y + \rho (u^+ - z^+) \label{eq:dual_update}
\end{align}
\end{subequations}
where the last step is a gradient-ascent update on the duals. These steps can be iterated until a desired convergence tolerance is achieved. Importantly, the primal problem~\ref{eq:primal_update} now reduces to the Riccati Recursion, the slack update is a projection, and the dual update is simply a vector addition.

The key observation that makes this whole framework efficient is that many common convex constraint sets admit simple, closed-form projection operators to solve~\eqref{eq:slack_update}.

Cache-based solvers, like TinyMPC~\cite{tinympc}, go one step further and fix and cache key values offline that reduce the computational complexity of the primal Riccati recursion even further. These computational savings both reduce the latency of the solver as well as its memory requirements, enabling high-rate MPC on resource-constrained platforms.


In particular, for sufficiently long horizons, the Riccati recursion converges to the infinite-horizon LQR solution~\cite{lewis12optimal}, where the time-varying matrices $K_k$ and $P_k$ are approximated by steady-state quantities $K_\infty$ and $P_\infty$. 

Combined with assumptions of fixed $A,B,Q,R,Q_f$, this simplifies \eqref{eq:lqr_backward} to:
\begin{equation}\label{eq:fast_riccati}
    \begin{aligned}
        d_k &= C_1(B^\top p_{k+1} + r_k), \\[-0.25em]
        p_k &= q_k + C_2 p_{k+1} - K_\infty^\top r_k,
    \end{aligned}
\end{equation}
where, $C_1 = (R + B^\top P_\infty B)^{-1}, C_2 = (A - BK_\infty)^\top$.


\subsection{Semidefinite Relaxations of Quadratic Constraints}

Many geometric safety constraints encountered in robotics are quadratic and
nonconvex. For clarity, we first consider planar obstacle avoidance whose constraints take the form (where $p_k \in \mathbb{R}^2$ denotes the position component of the state):

\begin{equation}
\|p_k - c_j\|_2^2 \ge r_j^2 ,
\label{eq:obstacle-constraints}
\end{equation}
where $c_j \in \mathbb{R}^2$ is the center and $r_j > 0$ the radius of obstacle $j$. Note that Equation~\eqref{eq:obstacle-constraints} is nonconvex.\footnote{The same lifting idea extends directly to a $d$-dimensional geometric subspace $p_k \in \mathbb{R}^d$, with disks replaced by Euclidean balls and the lifted moment block growing accordingly.}

In this paper, we adopt the standard semidefinite lifting of the geometric state. Specifically, we introduce the lifted matrix, 
\begin{equation}\label{eq:lift}
P_k = p_k p_k^\top ,
\end{equation}
which allows quadratic terms in $p_k$ to be expressed linearly. Consistency between
$p_k$ and $P_k$ is relaxed via the positive semidefinite constraint, equivalent to~\eqref{eq:lift} when~\eqref{eq:schur} is rank-1:
\begin{equation}\label{eq:schur}
\begin{bmatrix}
1 & p_k^\top \\
p_k & P_k
\end{bmatrix} \succeq 0.
\end{equation}
Such semidefinite relaxations are widely used in offline trajectory optimization, sum-of-squares programming, and safety verification~\cite{vandenberghe1996semidefinite, parrilo2005exploiting, henrion2005positive, majumdar2017funnel, marcucci2023motion, dong2026fast, papalia2025certifiable}. However, general-purpose semidefinite programming (SDP) solvers scale poorly with problem dimension and horizon length, making them unsuitable for real-time control and embedded deployment.

\section{TinySDP}
\label{sec:method}
Despite their success in embedded settings, Riccati-based MPC solvers are typically limited to constraint classes that preserve stage-wise structure and admit efficient proximal/projection steps (e.g., linear and second-order cone constraints). While semidefinite programming is also a form of conic optimization, it is considered incompatible with real-time MPC as general-purpose SDP methods require repeated large matrix factorizations with per-iteration costs that scale cubically in the semidefinite matrix variable dimensions.

We show that, by introducing a structured lifting that preserves the computational structure needed for efficient Riccati-based MPC, these tractability barriers can be bypassed, enabling real-time, certifiable safety for embedded robotics.

We consider the MPC problem~\eqref{eq:mpc} with the following inequality constraints. First, inputs are subject to box constraints:
\begin{equation} \label{eq:input_bounds}
    u_{\min} \le u_k \le u_{\max}.
\end{equation}
Second, safety is enforced via obstacle avoidance constraints on a selected geometric subspace of the state. In our deployed setting, we specialize to the
planar position and let
$p_k \in \mathbb{R}^2$ denote the planar position component of the state, i.e.,
$p_k = C_p x_k$ for a known selection matrix $C_p \in \mathbb{R}^{2 \times n_x}$.
We model obstacles as unions of disks,
\begin{equation} \label{eq:disks}
  \mathcal{O}_{k,j} := \{ p \in \mathbb{R}^2 : \|p - c_{k,j}\|_2^2 \le r_{k,j}^2 \},
\end{equation}
and enforce safety via per-timestep keep-out constraints using~\eqref{eq:obstacle-constraints}.
Such constraints are inherently nonconvex and present a central challenge for real-time MPC.
Our goal is therefore to design a receding-horizon MPC controller that enforces
\eqref{eq:obstacle-constraints} with explicit safety guarantees, while remaining compatible with embedded solvers based on Riccati recursion.

\subsection{PSD-Relaxed Lifted MPC}
\label{sec:lifted}

\subsubsection{Lifted Dynamics and Costs}
Applying the semidefinite lifting~\eqref{eq:lift} to ~\eqref{eq:schur}, we introduce auxiliary lifted second-order moment variables for the state and input outer products: 

\begin{equation*}
X_k = x_k x_k^\top,~~
XU_k = x_k u_k^\top,~~
UX_k = u_k x_k^\top,~~
UU_k = u_k u_k^\top,
\end{equation*}
which we lift to SDP constraints analogously to~\eqref{eq:schur}. We define the lifted state, input, and initial condition as,
\begin{equation*}
\resizebox{0.99\columnwidth}{!}{$
  \bar x_k :=
  \begin{bmatrix}
    x_k \\
    \mathrm{vec}(X_k)
  \end{bmatrix},
  \;
  \bar u_k :=
  \begin{bmatrix}
    u_k \\
    \mathrm{vec}(XU_k) \\
    \mathrm{vec}(UX_k) \\
    \mathrm{vec}(UU_k)
  \end{bmatrix},
  \;
  \bar x_0 =
  \begin{bmatrix}
    x_0 \\
    \mathrm{vec}(x_0x_0^\top)
  \end{bmatrix},
$}
\end{equation*}

noting that the lifted initial state is rank-consistent at initialization (as required by the rank-1 certificate introduced later).

Using the following identities and expansion,
\begin{align*}
\mathrm{vec}(Axx^\top A^\top) = (A \otimes A)\,\mathrm{vec}(xx^\top), \\[-0.25em]
\mathrm{vec}(Axu^\top B^\top) = (B \otimes A)\,\mathrm{vec}(xu^\top), \\[-0.25em]
x_{k+1}x_{k+1}^\top = (Ax_k + Bu_k)(Ax_k + Bu_k)^\top,
\end{align*}
the lifted dynamics induced by $x_{k+1} = Ax_k + Bu_k$ remain linear in the lifted variables ($\otimes$ is the Kronecker product):
\begin{equation}
  \bar x_{k+1} = \bar A\,\bar x_k + \bar B\,\bar u_k,
  \label{eq:lifted_dynamics}
\end{equation}

\begin{equation*}
\scalebox{0.93}{$
\bar A = 
\begin{bmatrix}
A & 0\\
0 & A\otimes A
\end{bmatrix},\;
\bar B =
\begin{bmatrix}
B & 0 & 0 & 0\\
0 & B\otimes A & A\otimes B & B\otimes B
\end{bmatrix}$}.
\end{equation*}
We can then define the projection matrices,
\[
C_x := \begin{bmatrix} I_{n_x} & 0 \end{bmatrix} ~~\mathrm{and}~~ C_u := \begin{bmatrix} I_{n_u} & 0 & 0 & 0 \end{bmatrix},
\]
enabling us to relate $x_k$ and $u_k$ to $\bar x_k$ and $\bar u_k$ via $x_k = C_x \bar x_k$ and $u_k = C_u \bar u_k$ respectively. The quadratic stage cost in~\eqref{eq:mpc} is then applied to the physical variables extracted from the lifted decision variables (with $\bar Q := C_x^\top Q C_x$ and $\bar R := C_u^\top R C_u$),
\begin{align}\label{eq:lifted_cost_selectors}
\ell(x_k,u_k) &= \bar x_k^\top \bar Q\,\bar x_k + \bar u_k^\top \bar R\,\bar u_k.
\end{align}

\subsubsection{Lifted Geometric Constraints}

As $p_k = C_p x_k \in \mathbb{R}^2$, we can define the corresponding block of the lifted second moment matrix as follows, noting that in the exact case where $X_k = x_k x_k^\top$ this reduces to $X_k^{(p)} = p_k p_k^\top$,
\begin{equation}
  X_k^{(p)} := C_p X_k C_p^\top \in \mathbb{R}^{2 \times 2}.
\end{equation}

While we specialize to $p_k \in \mathbb{R}^2$ for the planar obstacle-avoidance
setting studied in this paper, the same construction applies to any
$d$-dimensional geometric subspace by taking
$X_k^{(p)} := C_p X_k C_p^\top \in \mathbb{R}^{d \times d}$ and replacing disks
with Euclidean balls defined over the corresponding geometric coordinates.

The nonconvex disk obstacle avoidance constraints~\eqref{eq:disks} can also thus be expanded as,
\begin{equation*}
  \|p_k\|_2^2 - 2 c_{k,j}^\top p_k + \|c_{k,j}\|_2^2 \ge r_{k,j}^2 .
  \label{eq:obstacle_constraints_expansion}
\end{equation*}
Replacing $\|p_k\|_2^2$ with $\mathrm{trace}(X_k^{(p)})$ yields the following lifted affine inequality constraint which is linear in the lifted variables and is exact when $X_k = x_k x_k^\top$,
\begin{equation}
  \mathrm{trace}(X_k^{(p)})
  - 2 c_{k,j}^\top p_k
  + \|c_{k,j}\|_2^2
  \ge r_{k,j}^2.
  \label{eq:lifted-disk}
\end{equation}

\subsubsection{Per-Stage PSD Constraint}

We impose a per-stage positive semidefinite constraint on the augmented state--input moment matrix:
\begin{equation}\label{eq:psd-stage}
  M_k(\bar x_k,\bar u_k) :=
  \begin{bmatrix}
    1   & x_k^\top & u_k^\top \\
    x_k & X_k      & XU_k \\
    u_k & UX_k     & UU_k
  \end{bmatrix}
  \in \mathbb{S}^{1+n_x+n_u}.
\end{equation}

In the exact (non-relaxed) case, $M_k$ is rank-1. Relaxing to $M_k \succeq 0$ yields a convex relaxation that permits $X_k \neq x_k x_k^\top$ (and similarly for $XU_k$, $UX_k$, $UU_k$) when obstacle constraints are active. In our approach, the tightness of the relaxation is assessed a posteriori using the certificate in Section~\ref{sec:certificate}. The physical variables $x_k$ and $u_k$ remain explicit decision variables in the MPC and are not reconstructed from the lifted moment.

\subsubsection{Lifted MPC Problem}

The resulting MPC problem is posed over lifted variables $\{\bar x_k\}_{k=0}^N$ and $\{\bar u_k\}_{k=0}^{N-1}$ with:
(i) linear lifted dynamics~\eqref{eq:lifted_dynamics},
(ii) lifted costs~\eqref{eq:lifted_cost_selectors},
(iii) input bounds~\eqref{eq:input_bounds} on the physical inputs $u_k$,
(iv) lifted geometric inequalities~\eqref{eq:lifted-disk}, and
(v) per-stage PSD constraints~\eqref{eq:psd-stage}.
This problem is convex but includes per-stage semidefinite constraints, which are not directly supported by existing embedded solvers. We next show how to solve this problem efficiently using ADMM while preserving Riccati structure.

\subsection{Riccati--ADMM Solver for PSD-Relaxed Lifted MPC}
\label{sec:admm}

We solve the PSD-relaxed lifted MPC problem using a Riccati-based ADMM, extending the TinyMPC framework~\cite{tinympc} to incorporate per-stage semidefinite constraints while preserving its computational structure.

We introduce a PSD slack variable $S_k \in \mathbb{S}_+^{p}$ and a scaled dual
variable $H_k \in \mathbb{S}^{p}$ for the coupling constraint, which are both stored using the $\sqrt{2}$-scaled half-vectorization operator $\mathrm{svec}(\cdot)$, with inverse mapping $\mathrm{smat}(\cdot)$, to further reduce the memory requirements of our approach, while ensuring that Frobenius inner products are preserved under vectorization.
We then define $M_k(\bar x_k,\bar u_k)$ as the per-stage state--input moment matrix defined in
\eqref{eq:psd-stage} and $p := 1+n_x+n_u$,
\begin{equation}
  M_k(\bar x_k,\bar u_k) = S_k.
\end{equation}

Using the scaled form of ADMM, each iteration alternates between a primal update,
a PSD projection, and a dual update. To simplify notation, we define the augmented
stage cost,
\begin{equation} \label{eq:aug_stage_cost}
\underbrace{\ell_k(\bar x_k, u_k)
+
\frac{\rho_{\mathrm{psd}}}{2}
\big\| M_k(\bar x_k,\bar u_k) - S_k + H_k \big\|_F^2}_{\tilde{\ell}_k(\bar x_k, u_k; S_k, H_k)}.
\end{equation}

The ADMM updates are thus given by,
\begin{subequations}
\begin{align}
  (\bar x, u)^+
  &=
  \argmin_{\bar x, u}
  \sum_{k=0}^{N-1}
  \tilde{\ell}_k(\bar x_k, u_k; S_k, H_k),
  \label{eq:admm_primal}
  \\
  S_k^+
  &=
\Pi_{\mathbb{S}_+^{p}}
\!\big( M_k(\bar x_k^+,\bar u_k^+) + H_k \big),
  \label{eq:admm_psdproj}
  \\
  H_k^+
  &=
  H_k
  +
  \gamma_{\mathrm{psd}}
  \big( M_k(\bar x_k^+,\bar u_k^+) - S_k^+ \big).
\label{eq:admm_dual}
\end{align}
\end{subequations}
where $\rho_{\mathrm{psd}} > 0$ is the penalty parameter, and $\gamma_{\mathrm{psd}} \in (0,1]$ is an optional under-relaxation factor.
Here $\Pi_{\mathbb{S}_+^{p}}(\cdot)$ denotes the Euclidean projection onto the positive semidefinite cone:
\begin{equation*}
\Pi_{\mathbb{S}_+^{p}}(Y) := \arg\min_{X \succeq 0}\ \|X - Y\|_F^2 .
\end{equation*}
Note that this projection has an analytic solution,
\begin{equation}\label{eq:thresh}
\Pi_{\mathbb{S}_+^{p}}(Y)=V\operatorname{diag}\!\big(\max(\lambda,0)\big)V^\top,
\end{equation}
where $V\operatorname{diag}(\lambda)V^\top$ is the eigen-decomposition of $Y$.

\subsubsection{Primal Update (Riccati Step)}

With the slack and dual variables fixed, the primal subproblem~\eqref{eq:admm_primal}
is an equality-constrained quadratic program with linear dynamics and the lifted
quadratic stage cost defined in~\eqref{eq:lifted_cost_selectors}.
The PSD-augmented term in Eq.~\eqref{eq:aug_stage_cost} can be expanded as follows, where $\|\cdot\|_F$ and $\langle \cdot,\cdot\rangle_F$ denote the Frobenius norm and Frobenius inner product, respectively, and we define $T_k := S_k - H_k$:
\begin{align*}
\frac{\rho_{\mathrm{psd}}}{2}
\| M_k - S_k + H_k \|_F^2
&=
\frac{\rho_{\mathrm{psd}}}{2}\|M_k\|_F^2 \\
&\quad
-\rho_{\mathrm{psd}}\langle T_k,M_k\rangle_F
+\text{constant}.
\end{align*}

Since the $M_k$ depends affinely on the lifted
variables, the inner-product term contributes additional linear terms in
$\bar x_k$ and $\bar u_k$. Thus, for fixed $(S_k,H_k)$, the primal stage cost
retains the form
$
\bar x_k^\top \bar Q\,\bar x_k
+
\bar u_k^\top \bar R\,\bar u_k
+
q_k^\top \bar x_k
+
r_k^\top \bar u_k,
$
where the linear coefficients $q_k$ and $r_k$ incorporate the adjoint pullback of
the PSD term. This corresponds to the ``update linear costs via adjoint mapping''
step in Algorithm~\ref{alg:tinysdp}.

Crucially, the system dynamics~\eqref{eq:lifted_dynamics} remain unchanged, so the resulting primal problem is still a lifted LQR problem. As in TinyMPC~\cite{tinympc}, it can therefore be solved efficiently by a backward-forward Riccati sweep using the cached steady-state Riccati quantities computed offline (as described in Section~\ref{sec:background}).

\subsubsection{PSD Projection and Dual Update}

At each ADMM iteration, the slack update~\eqref{eq:admm_psdproj} is computed via~\eqref{eq:thresh}, followed by the scaled dual update~\eqref{eq:admm_dual}. 

\begin{algorithm}[t]
\caption{TinySDP: PSD-Relaxed Lifted MPC}\label{alg:tinysdp}
\begin{algorithmic}[1]
\Function{TinySDP\_Offline}{Lifted Model \& Costs}
    \State Construct lifted dynamics $(\bar{A}, \bar{B})$ via \eqref{eq:lifted_dynamics}
    \State Construct lifted quadratic costs $(\bar Q,\bar R)$ via \eqref{eq:lifted_cost_selectors}
        \State Compute and cache lifted Riccati quantities via \eqref{eq:lqr_backward}
   \State \Return Cached solver data $\mathcal{C}$
\EndFunction

\Function{TinySDP\_Online}{$x_k$, Obstacles $\mathcal{O}_k$, Cache $\mathcal{C}$}
\State Lift current state: $\bar x_0 \gets [x_k^\top, \mathrm{vec}(x_k x_k^\top)^\top]^\top$

    \State Initialize slack and dual variables
    
    \While{residual $> \varepsilon$ and iter $< K_{\max}$}
    
        \State \textcolor{blue}{\texttt{// Primal Update (Riccati Step)}}
        \State $q_k, r_k \gets$ Update linear costs via adjoint mapping
        \State $d_k, p_k \gets$ Backward Riccati recursion via \eqref{eq:fast_riccati}
        \State $\bar{x}, \bar{u} \gets$ Forward rollout via \eqref{eq:lqr_forward}

        \State \textcolor{blue}{\texttt{// PSD Projection and Dual Update}}
        \For{$k = 0$ to $N-1$}
          \State Form moment matrix $M_k(\bar{x}_k,\bar{u}_k)$ via \eqref{eq:psd-stage}
          \State $S_k \gets$ PSD Projection via \eqref{eq:admm_psdproj}
          \State $H_k \gets$ Dual Update via \eqref{eq:admm_dual}
       \EndFor
    \EndWhile
    \State \Return $(\bar x_{k:k+N}, \bar u_{k:k+N-1})$
\EndFunction
\end{algorithmic}
\end{algorithm}
\begin{algorithm}[t]
\caption{TinySDP with Online Certificate}\label{alg:tinysdp_mpc_certificate}
\begin{algorithmic}[1]
\Function{TinySDP\_Step}{$x_k,\mathcal{O}_k,\mathcal{C},u_{\mathrm{hover}}$}

    \State \textcolor{blue}{\texttt{// Solve with TinySDP}}
    \State $\bar{x}_{k:k+N},\bar{u}_{k:k+N-1} \gets \textsc{TinySDP\_Online}(x_k,\mathcal{O}_k,\mathcal{C})$

    \State \textcolor{blue}{\texttt{// Verify solution quality}}
    \State $\Delta_k \gets$ Compute trace gap via \eqref{eq:trace_gap}
    \State $\eta_k^{\min} \gets \min_j \eta_{k,j}$ via \eqref{eq:lifted_margin}

    \If{$\eta_k^{\min} \ge 0$ \textbf{and} $|\Delta_k| \le \eta_k^{\min}$}
        \State $u_k \gets \bar{u}_{k}$ \Comment{First control in optimized sequence}

    \Else
        \State $u_k \gets u_{\mathrm{hover}}$ \Comment{Fallback Policy (brake and hover)}
    \EndIf
    \State \Return $u_k$
\EndFunction
\end{algorithmic}
\end{algorithm}

\subsection{Overall TinySDP Algorithm}
TinySDP (Algorithm~\ref{alg:tinysdp}) adapts a standard receding-horizon MPC loop. Offline, the lifted dynamics, costs, and steady-state Riccati quantities for the lifted primal subproblem are computed and stored in a cache, $\mathcal C := \{K_\infty, P_\infty, C_1, C_2, \ldots\}$. Online, at each control timestep $k$, the current measured state $x_k$ and obstacle set $\mathcal{O}_k$ are used to solve the finite-horizon PSD-relaxed lifted MPC problem. Within each ADMM iteration, the cached infinite-horizon Riccati quantities are used for efficiency. The solver returns a finite-horizon state/input sequence. The first control input is then applied and the problem is re-solved at the next control step using updated state and obstacle information. 

\subsection{A Posteriori Rank-1 Safety Certificate}
\label{sec:certificate}

\paragraph{Trace Gap and Lifted Margin}
For clarity, we state the certificate for the planar case $p_k \in \mathbb{R}^2$, but the same argument extends directly to any $d$-dimensional geometric subspace by replacing $X_k^{(p)} \in \mathbb{R}^{2\times2}$ with $X_k^{(p)} \in \mathbb{R}^{d\times d}$.
Let $p_k \in \mathbb{R}^2$ be the position at time $k$ and $X_k^{(p)} \in \mathbb{R}^{2 \times 2}$ be the position block of the lifted moment matrix. We define the \emph{trace gap} as
\begin{equation} \label{eq:trace_gap}
    \Delta_k := \mathrm{trace}(X_k^{(p)}) - \|p_k\|_2^2.
\end{equation}
This scalar measures the mismatch between the relaxed lifted variable and the exact rank-1 moment induced by the physical state $p_k$. In particular, $\Delta_k = 0$ implies that the lifted position block is consistent with the exact physical second moment, while $\Delta_k > 0$ indicates that the relaxed lifted second moment occupies more volume than the physical state. Similarly, for obstacle $j$ with center $c_j$ and radius $r_j$, the \emph{lifted margin} is
\begin{equation} \label{eq:lifted_margin}
    \eta_{k,j} := \mathrm{trace}(X_k^{(p)}) - 2 c_j^\top p_k + \|c_j\|_2^2 - r_j^2,
\end{equation}
which corresponds to the linear obstacle-avoidance constraint enforced by the solver.

\paragraph{Certification Logic}
The true squared clearance to obstacle $j$ is given by $\delta_{k,j} := \|p_k - c_j\|_2^2 - r_j^2$. Substituting the definitions above yields the identity:
$
    \delta_{k,j} = \eta_{k,j} - \Delta_k.
$
Let $\eta_k^{\min} := \min_j \eta_{k,j}$. It follows that if,
\begin{equation} \label{eq:cert_condition}
    \eta_k^{\min} \ge 0 \quad \text{and} \quad |\Delta_k| \le \eta_k^{\min},
\end{equation}
then $\delta_{k,j} \ge 0$ for all $j$, certifying that $x_k$ is collision-free.

The rank-1 condition is \emph{sufficient}, but not necessary. A certification failure occurs when either the lifted margin is negative ($\eta_k^{\min} < 0$) or the relaxation gap exceeds the available lifted margin ($|\Delta_k| > \eta_k^{\min}$). However, we note that, even when the relaxation is not exactly rank-1, this simple trace-based certificate~\eqref{eq:cert_condition} suffices to certify geometric clearance from obstacles. When the certificate fails, the relaxation may be too loose, and while the physical state $p_k$ might still be safe, we cannot provide a mathematical guarantee.

\paragraph{Online Safety Monitor}

To enhance robustness in deployed settings, we integrate a runtime safety monitor into the control loop (Algorithm~\ref{alg:tinysdp_mpc_certificate}). If the solver produces a solution that fails~\eqref{eq:cert_condition}, the system discards the planned update and executes a fallback stop-and-hover policy. This ensures that uncertified control inputs are never applied to the physical system. While this monitor provides a rigorous check on the deployed state, it serves as a reactive failsafe rather than a formal closed-loop guarantee under extreme conditions, such as high-velocity obstacles or significant modeling errors.

\section{Experiments}
\label{sec:experiments}

We designed our experiments to evaluate both static and dynamic obstacle avoidance under known challenging geometries for state-of-the-art methods. In the static setting, we consider a concave U-shaped obstacle, a cul-de-sac, with the goal located beyond the obstacle, a canonical cause of deadlock and failure modes for local and reactive methods~\cite{fox2002dynamic,rajagopal2025dr}. In the dynamic setting, we construct a scenario consisting of a static obstacle directly in front of the agent together with two moving obstacles that periodically close a narrow passage, resulting in a three-obstacle interaction that requires continuous updates to the obstacle avoidance strategy, rather than one-shot reactive planning. We finally deploy our approach on a Crazyflie to evaluate our method's real-world feasibility.  

We compare our TinySDP method against three baselines: \begin{itemize}
    \item TinyMPC-LIN: linearized tangent half-space constraints as per the original TinyMPC paper~\cite{tinympc},
    \item TinyMPC-HOCBF: relative-degree-2 control barrier function constraints with TinyMPC, inspired by~\cite{xiao2021high, zeng2021safety},
    \item RPCBF~\cite{knoedler2025safety}, a state-of-the-art sampling-based policy safety filter implemented outside TinyMPC.
\end{itemize} 

For all our experiments, all methods use identical discrete-time dynamics, input bounds, horizon length, and obstacle geometry, unless otherwise specified. Additionally, to ensure a fair comparison, we evaluate TinyMPC-LIN and TinyMPC-HOCBF with a zero safety margin, benchmarking against TinySDP which inherently requires no additional margin. For RPCBF, we similarly use zero safety margin but tune the underlying $\alpha$ parameters to achieve the best possible performance. 
Additionally, to provide a ``best-case'' analysis for the linearized baselines, we performed a grid search to determine the minimum safety margin required for TinyMPC-LIN and TinyMPC-HOCBF to successfully navigate the benchmarks. Our ablation studies reveal that these methods require a minimum safety margin of $1.5\,\mathrm{m}$ to avoid collision. We note that such margins are highly impractical for our target domain of embedded, small-scale robotic platforms, which have widths roughly equal to $100\,\mathrm{mm}$~\cite{giernacki2017crazyflie}, resulting in artificial inflation of the collision footprint by over $17\times$. 


\subsection{Simulation: Static U-Shape Cul-de-Sac}

\begin{figure}[t!]
  \centering
  \includegraphics[width=0.9\columnwidth]{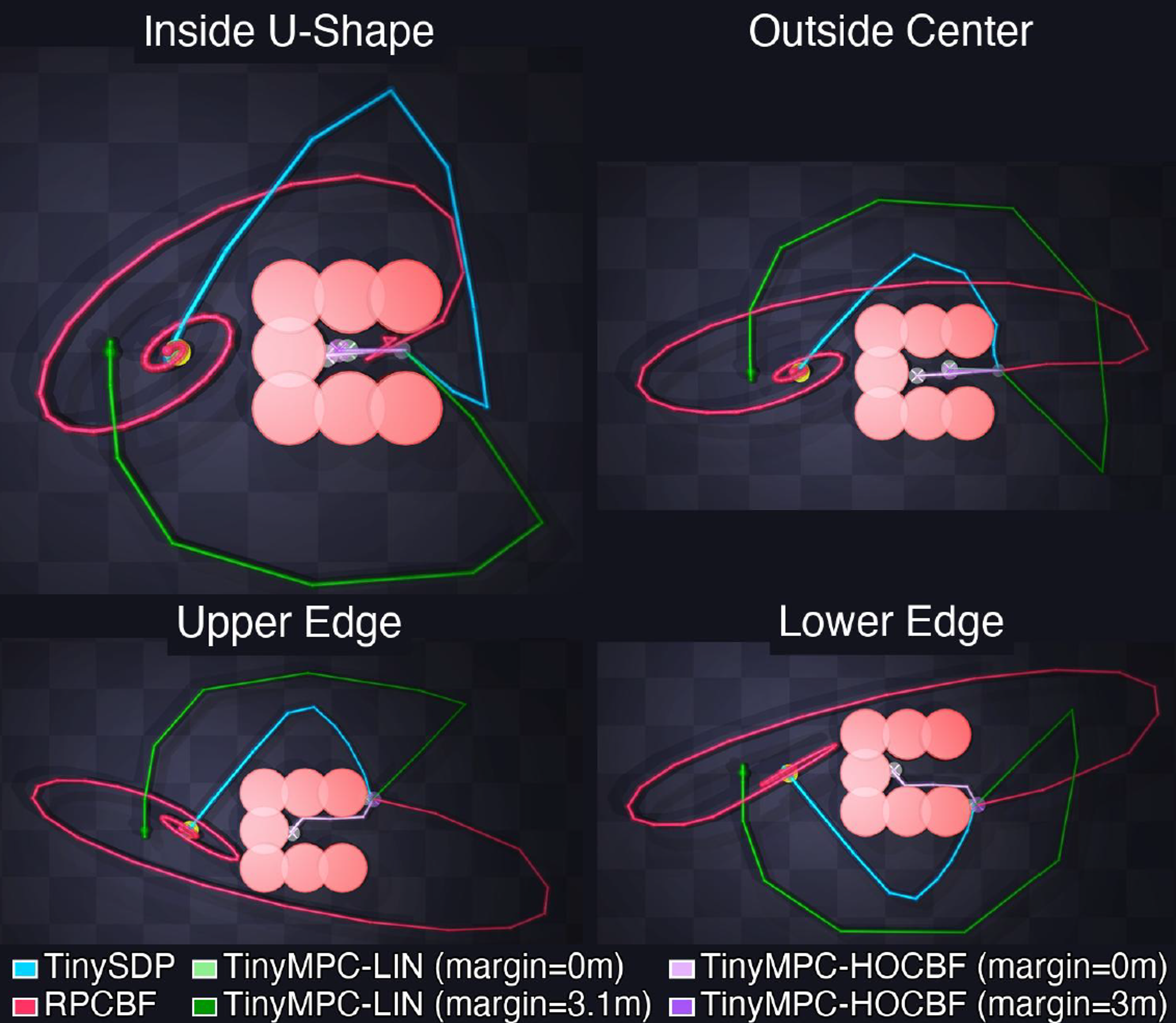}
  \vspace{-3pt}
  \caption{\textbf{Static U-shape benchmark.} Trajectories for TinySDP (blue), RPCBF (red), TinyMPC-LIN (green), and TinyMPC-HOCBF (purple) across four initial conditions. 
  TinySDP consistently navigates the cul-de-sac to reach the goal. 
  TinyMPC-LIN (light green) and TinyMPC-HOCBF with any margin crash or get stuck. 
  While the tuned TinyMPC-LIN (dark green) becomes safe with a $3.1\,\mathrm{m}$ margin, this inflated buffer forces the solver to terminate far from the actual goal. RPCBF is able to safely navigate the challenging scenario, but yields significantly longer, more conservative paths than TinySDP.}
  \label{fig:ushape_grid}
  \vspace{-10pt}
\end{figure}

We first evaluate TinySDP against the baseline methods using a static U-shaped obstacle composed of overlapping disks.
Four representative initial conditions are considered: starting inside the cul-de-sac, outside the center opening, near the upper edge, and near the lower edge. 
Figure~\ref{fig:ushape_grid} visualizes the resulting trajectories, and Table~\ref{tab:ushape} summarizes performance across path length, distance to goal, and collision status.

TinyMPC-LIN and TinyMPC-HOCBF with a safety margin of $0$ collide for every start scenario.
Failures in TinyMPC-LIN arise from its reliance on nominal-dependent tangent half-space constraints, which do not impose a global geometric keep-out condition and may remain inactive until the system is already in an inevitable collision state.
Similarly, TinyMPC-HOCBF enforces a continuous-time barrier condition evaluated only at discrete sampling instants; in the presence of bounded inputs and concave, trap-like geometries, this local reactive formulation fails to prevent collisions.

While TinyMPC-LIN can navigate the cul-de-sac with a safety margin of $3.1\,\mathrm{m}$, this buffer is approximately $30\times$ the Crazyflie diameter, rendering the approach impractical for real-world deployment.
Moreover, the inflated obstacle boundaries overlap with the goal region, forcing the solver to terminate approximately $1.4\,\mathrm{m}$ away from the target.
TinyMPC-HOCBF fails to avoid collisions \emph{regardless of the safety margin}, highlighting the limitations of local reactive methods.

RPCBF~\cite{knoedler2025safety} is the strongest baseline and also remains collision-free in all cases with a 0m margin. However, it produces significantly longer, conservative trajectories.

Overall, TinySDP achieves $31\text{--}73\%$ shorter paths than RPCBF and $2\text{--}59\%$ shorter paths than the tuned TinyMPC-LIN.
TinySDP also consistently reaches within $0.02\,\mathrm{m}$ of the goal across all scenarios, making it $4\text{--}15\times$ more precise than RPCBF and $70\times$ more precise than TinyMPC-LIN.

\begin{table}[t!]
\centering
\small
\setlength{\tabcolsep}{3pt}
\caption{Static U-shape results for four initial conditions.}
\vspace{-5pt}
\label{tab:ushape}
\begin{tabular}{llccc}
\toprule
Start & Method & Path Len & Goal Dist & Safe \\
\midrule
Inside & TinySDP (ours) & \textbf{17.95} & \textbf{0.006} & \cmark \\
& RPCBF & 26.03 & 0.091 & \cmark \\
& TinyMPC-LIN (m=3.1m) & 18.38 & 1.400 & \cmark \\
& TinyMPC-LIN (m=0m) & -- & -- & \xmark \\
& TinyMPC-HOCBF (m=3m) & -- & -- & \xmark \\
& TinyMPC-HOCBF (m=0m) & -- & -- & \xmark \\
\midrule
Outside & TinySDP (ours) & \textbf{10.15} & \textbf{0.021} & \cmark \\
Center & RPCBF & 31.03 & 0.093 & \cmark \\
& TinyMPC-LIN (m=3.1m) & 24.58 & 1.400 & \cmark \\
& TinyMPC-LIN (m=0m) & -- & -- & \xmark \\
& TinyMPC-HOCBF (m=3m) & -- & -- & \xmark \\
& TinyMPC-HOCBF (m=0m) & -- & -- & \xmark \\
\midrule
Edge Up & TinySDP (ours) & \textbf{9.93} & \textbf{0.023} & \cmark \\
& RPCBF & 36.81 & 0.132 & \cmark \\
& TinyMPC-LIN (m=3.1m) & 18.58 & 1.400 & \cmark \\
& TinyMPC-LIN (m=0m) & -- & -- & \xmark \\
& TinyMPC-HOCBF (m=3m) & -- & -- & \xmark \\
& TinyMPC-HOCBF (m=0m) & -- & -- & \xmark \\
\midrule
Edge Down & TinySDP (ours) & \textbf{9.93} & \textbf{0.023} & \cmark \\
& RPCBF & 36.25 & 0.108 & \cmark \\
& TinyMPC-LIN (m=3.1m) & 23.64 & 1.400 & \cmark \\
& TinyMPC-LIN (m=0m) & -- & -- & \xmark \\
& TinyMPC-HOCBF (m=3m) & -- & -- & \xmark \\
& TinyMPC-HOCBF (m=0m) & -- & -- & \xmark \\
\bottomrule
\end{tabular}
\vspace{-5pt}
\end{table}


\subsection{Simulation: Dynamic Obstacles}

We next evaluate all methods in a dynamic obstacle scenario consisting of a static disk at the start, followed by moving disks that periodically open and close a narrow passage. This task requires continuous updates to the obstacle avoidance strategy, rather than one-shot reactive planning.

\begin{figure}[t!]
  \centering
  \includegraphics[width=0.9\columnwidth]{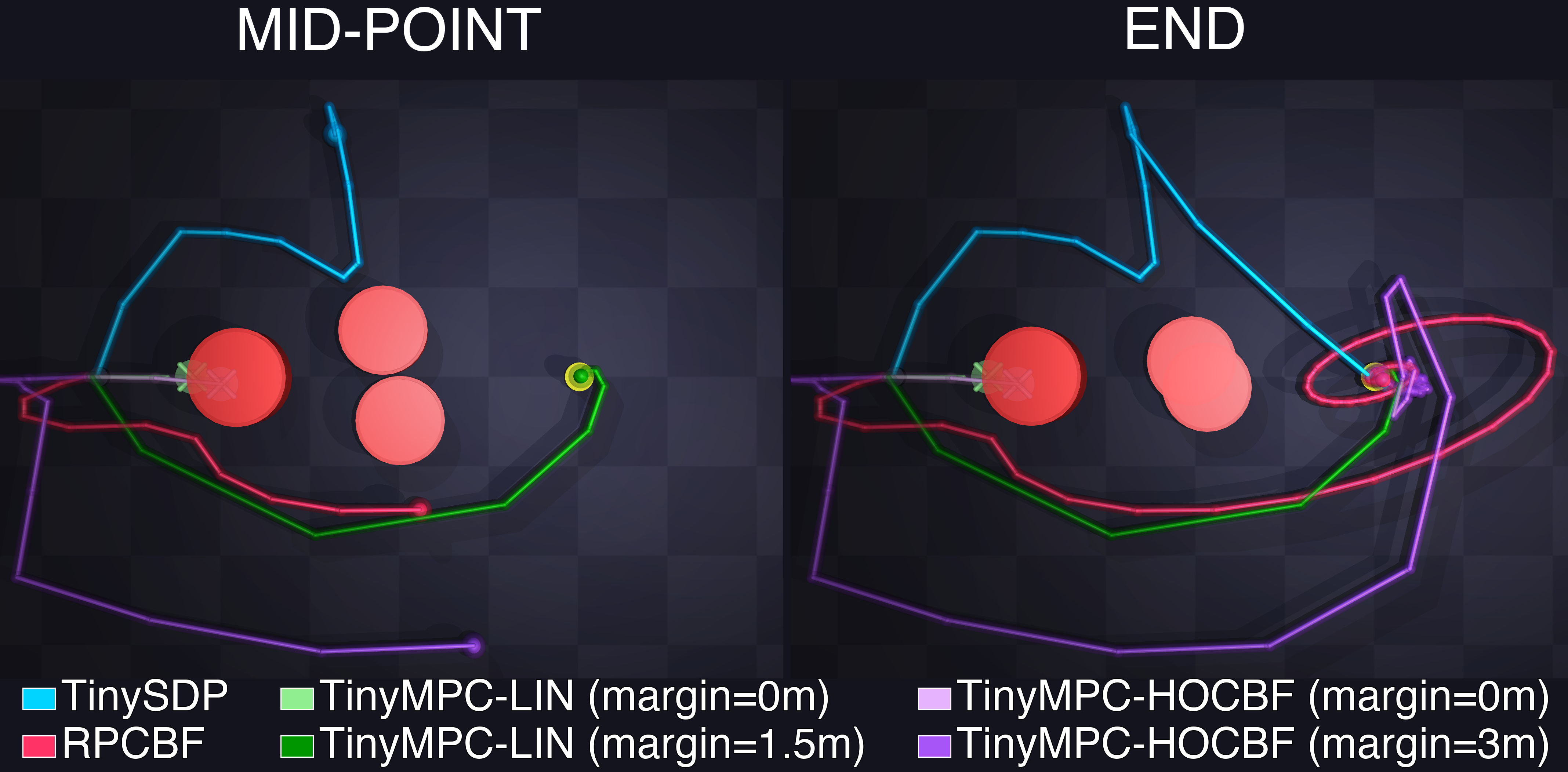}
  \vspace{-3pt}
  \caption{\textbf{Dynamic moving-gap benchmark.} Time-lapse comparison of TinySDP (blue) against RPCBF (red) and the linearized baselines (green, purple).
  TinySDP anticipates the moving obstacle, deviating early to maintain clearance. 
  The zero-margin baselines (light green, light purple) fail to anticipate the motion and collide. 
  The tuned baselines (dark green, dark purple) achieve safety only by using impractically large margins ($1.5\,\mathrm{m}$ and $3\,\mathrm{m}$ which are $17\text{--}30\times$ the drone's size). }
  \label{fig:dynamic_traj}
  \vspace{-10pt}
\end{figure}

Quantitative results are summarized in Table~\ref{tab:dynamic}, with representative trajectories shown in Figure~\ref{fig:dynamic_traj}. As illustrated, TinySDP successfully reaches the goal while maintaining positive clearance and remaining rank-1 at every timestep. In contrast, TinyMPC-LIN and TinyMPC-HOCBF with no safety margin both collide early, as local linearizations and discrete-time barrier evaluations are insufficient for fast-moving geometries. While these methods can be made safe using margins of $1.5\,\mathrm{m}$ and $3\,\mathrm{m}$ respectively, such values are $17\text{--}30\times$ the size of the drone and are therefore impractical for real dynamic environments with narrow corridors. RPCBF~\cite{knoedler2025safety} again remains collision-free with significantly longer, conservative paths.

Quantitatively, TinySDP achieves $30\%$ shorter paths than RPCBF and $43\%$ shorter paths than the tuned TinyMPC-HOCBF ($3\,\mathrm{m}$ margin). While the tuned TinyMPC-LIN ($1.5\,\mathrm{m}$ margin) produces the shortest path at $13.61\,\mathrm{m}$, TinySDP converges $1.3\times$ closer to the goal ($0.018\,\mathrm{m}$ vs $0.023\,\mathrm{m}$) and is $3.8\text{--}4.3\times$ more precise than RPCBF and TinyMPC-HOCBF, respectively. The shorter path length of the tuned TinyMPC-LIN could be attributed to its specific trajectory, which curves from below and arrives at the intersection only after the moving obstacle has passed. This timing allows the solver to effectively, and luckily, bypass the active avoidance maneuver. In contrast, TinySDP actively deviates to negotiate the obstacle while it is still blocking the path, ensuring certified safety at the cost of a slightly longer trajectory in this specific case.


\subsection{Simulation: Extension to 3D Obstacle Avoidance Demos}

As noted earlier, the proposed lifting is not inherently limited to planar obstacles. In a 3D setting, lifting the position state $[1,x,y,z]^\top$ yields a $4\times4$ moment matrix, so the same lifted semidefinite obstacle-avoidance formulation and applied-state certificate structure extend naturally to spherical obstacles and other quadratic sets (e.g., ellipsoids, cylinders).

Of course, these richer geometric formulations require larger PSD cones and therefore more expensive projections, inducing tighter embedded compute and memory tradeoffs. 

\begin{figure*}[!t]
  \centering

  \includegraphics[width=\textwidth]{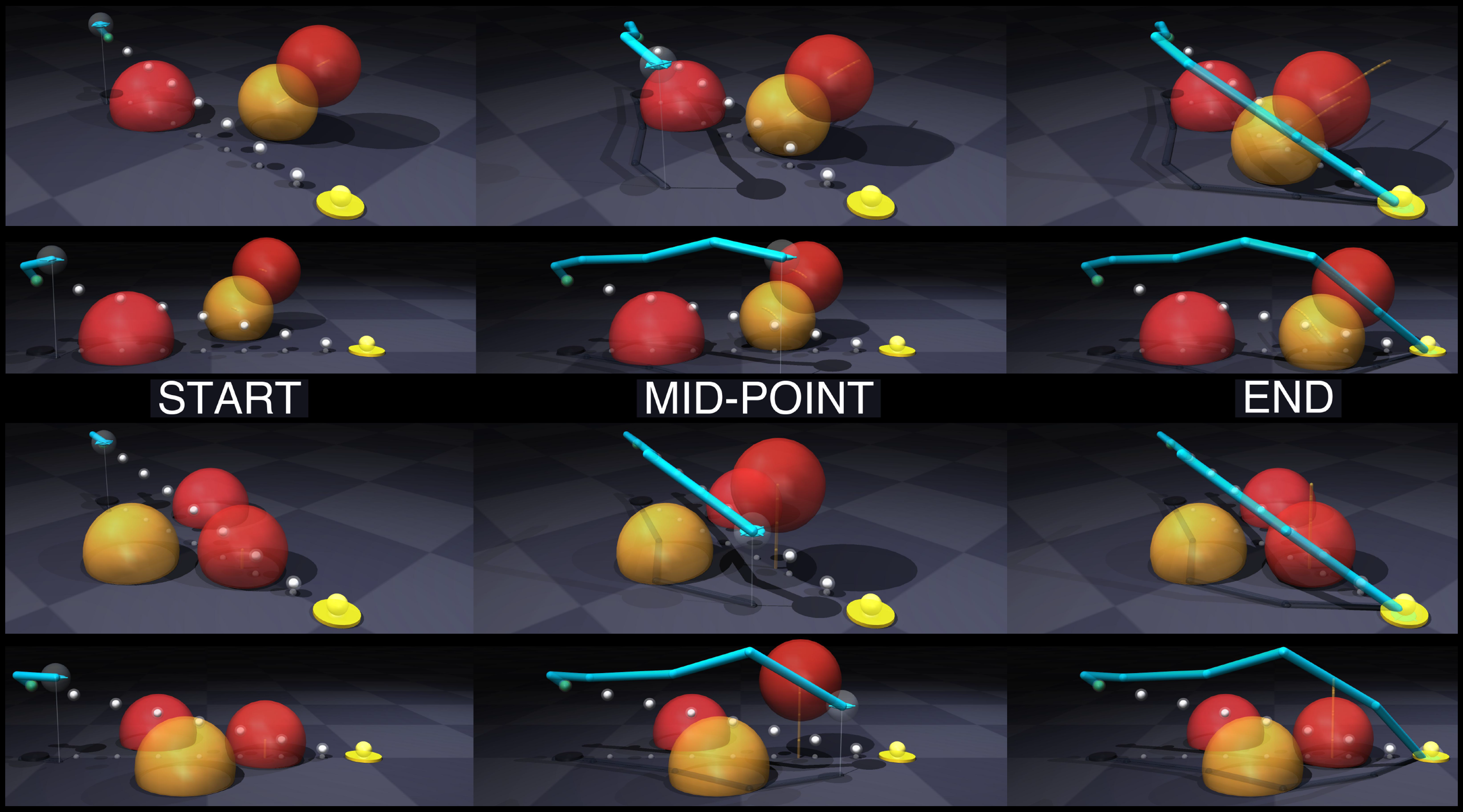}
  \vspace{-3pt}
  \caption{\textbf{3D dynamic obstacle avoidance extension.} TinySDP navigating two dynamic 3D sphere scenarios: \emph{Sweeping Barrier} (top) and \emph{Vertical Gate} (bottom). The two rows provide complementary views of the same 3D motion, while the columns show the approach, midpoint, and end of the maneuver. The white dotted path indicates the direct start-to-goal line, which would intersect the obstacle field. TinySDP extends naturally to 3D sphere avoidance, executing nonplanar maneuvers with positive clearance in both scenarios.}
  \label{fig:3d_extension}

\end{figure*}

To illustrate this extension, we consider two dynamic 3D scenarios. In \emph{Sweeping Barrier}, two moving spheres sweep laterally across the route while a third upstream sphere blocks the direct start-to-goal path, requiring an early nonplanar deviation rather than a purely reactive planar sidestep. In \emph{Vertical Gate}, two static side spheres together with a centrally located sphere oscillating in height create a time-varying 3D passage, forcing a timed ascent-descent maneuver rather than a planar bypass. Representative trajectories are shown in Figure~\ref{fig:3d_extension}. In both scenarios, TinySDP reaches the goal without collisions.

\begin{table}[!t]
\centering
\caption{Dynamic obstacle scenario results.}
\vspace{-5pt}
\label{tab:dynamic}
\begin{tabular}{lccc}
\toprule
Method & Path Len & Goal Dist & Safe \\
\midrule
TinySDP (ours) & 19.11 & \textbf{0.018} & \cmark \\
RPCBF & 27.33 & 0.069 & \cmark \\
TinyMPC-LIN (m=1.5m) & \textbf{13.61} & 0.023 & \cmark \\
TinyMPC-LIN (m=0m) & -- & -- & \xmark \\
TinyMPC-HOCBF (m=3m) & 33.80 & 0.077 & \cmark \\
TinyMPC-HOCBF (m=0m) & -- & -- & \xmark \\
\bottomrule
\end{tabular}
\vspace{-5pt}
\end{table}


\begin{figure}[!t]
  \centering
  \includegraphics[width=0.9\columnwidth]{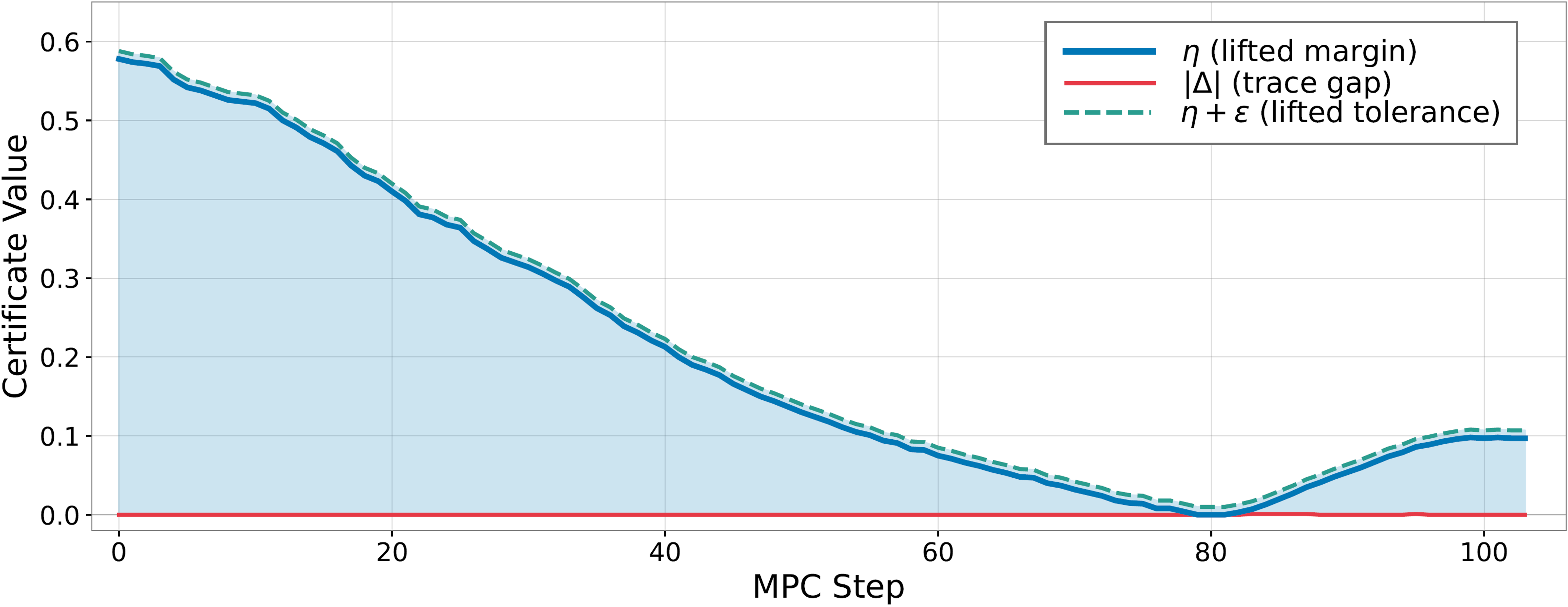}
  \vspace{-3pt}
  \caption{Evolution of the Rank-1 certificate as the Crazyflie avoided the moving arm, demonstrating certifiable safety (shaded region) during real-world operation.}
  \label{fig:rank1}
  \vspace{-10pt}
\end{figure}

\subsection{Real World Deployment: Dynamic Obstacles}

To validate real-world feasibility on resource-constrained platforms, we deploy TinySDP on a Crazyflie 2.1 Brushless quadrotor, powered by an STM32F405 microcontroller (168 MHz with only 192 KB of SRAM). We apply the semidefinite relaxation strictly to the planar position of the quadrotor's 12-dimensional state, $(p_x, p_y)$, yielding a $3\times3$ moment matrix,
\begin{equation*}
  M_k = \begin{bmatrix} 1 & p_x & p_y \\ p_x & p_x^2 & p_x p_y \\ p_y & p_x p_y & p_y^2 \end{bmatrix},
\end{equation*}
where the PSD constraint $M_k \succeq 0$ is enforced via projection within the ADMM solver. Since the nonconvex obstacle constraints depend only on planar position, this corresponds to the minimal task-induced lift sufficient to certify geometric clearance. The remaining state variables (altitude, velocity, attitude) are handled via box constraints as in prior work~\cite{tinympc}.

The solver runs onboard at 25~Hz with a 20-step horizon. In the deployed implementation, ADMM is capped at 5 iterations per control update to guarantee timing. In the latest logged hardware run, the solver executed all 5 iterations and reported a total MPC solve time of $14.842\,\mathrm{ms}$, leaving approximately $25.2\,\mathrm{ms}$ of margin within the $40\,\mathrm{ms}$ control period. The compiled firmware footprint is $272.4\,\mathrm{KB}$ of flash and $139.0\,\mathrm{KB}$ of on-chip RAM ($83.5\,\mathrm{KB}$ main SRAM and $55.6\,\mathrm{KB}$ CCM). The PSD-specific overhead is modest, adding approximately $1.47\,\mathrm{KB}$ of fixed static memory, about $720\,\mathrm{B}$ of peak additional stack usage along the PSD projection path, and an estimated $2.5$--$3.4\,\mathrm{KB}$ of flash. Since the lifted variable is only a symmetric $3\times3$ matrix over $[1,p_x,p_y]$, the PSD projection is implemented using a specialized small-matrix eigendecomposition with eigenvalue clamping and is not a practical bottleneck in this planar setting.

A low-level PID controller tracks the planned trajectory at 500~Hz. We validate the approach in a dynamic obstacle avoidance scenario where the Crazyflie flies 1~m forward while a robotic arm sweeps horizontally across its path. As shown in Figure~\ref{fig:hardware}, the drone successfully anticipates the moving obstacle, autonomously deviating to maintain safety before converging to the goal. Notably, as shown in Figure~\ref{fig:rank1}, the online safety certificate (Section~\ref{sec:certificate}) remained valid\footnotemark{} ($\Delta_k \approx 0$) at every control step throughout the flight for five independent trials, confirming that the rank-1 geometry was preserved even under real-world disturbances.
\footnotetext{
The certificate is evaluated with a small numerical tolerance
($\varepsilon = 10^{-2}$) to account for finite-precision arithmetic on the
embedded hardware platform.
}

\section{Limitations}
While TinySDP achieved zero failures across all hardware trials, several limitations warrant discussion. Most fundamentally, our trace-gap certificate provides a rigorous per-timestep geometric check on the applied state, but does not constitute a formal closed-loop guarantee. In particular, we do not prove recursive feasibility over the full predicted horizon, although offline analysis confirms horizon-wide certificate satisfaction on our nominal trajectories, this remains an empirical observation rather than a property of the closed-loop system itself.
While simulations show our lifting framework extends naturally to 3D spherical obstacles, onboard deployment reveals a computational tradeoff. The 3x3 moment matrix used in our planar hardware demonstration keeps PSD projections tractable; however, higher dimensions or complex geometries expand the lifted block, driving up per-iteration eigendecomposition costs. Transitioning this full 3D capability to microcontrollers remains a key focus for future work.
Finally, as established in our discussion of the online safety monitor~\ref{sec:certificate}, the fallback policy acts solely as a reactive failsafe. Extreme conditions such as sufficiently fast-moving obstacles, large modeling errors, or severe external disturbances could in principle outpace this fallback. Extending the framework with forward reachability analysis to provide proactive rather than purely reactive recovery guarantees is a natural next step.

\section{Conclusions and Future Work} 
We introduced TinySDP, a PSD-relaxed lifted MPC framework that enables certifiable obstacle avoidance on embedded platforms while preserving cached-Riccati-based computational structure.
By integrating per-stage semidefinite lifting with an ADMM-based solver, we establish a methodological pathway for incorporating semidefinite relaxations into embedded MPC. 
Furthermore, by exploiting per-stage structure and separating optimization from certification, we show that SDP-based safety constraints are not inherently incompatible with real-time receding-horizon control. 
This perspective bridges the gap between offline semidefinite planning and practical embedded deployment, and suggests that richer convex safety models can be integrated into high-rate controllers when solver design respects MPC structure. 
Future work includes extending the framework to more challenging hardware demonstrations and richer obstacle representations to identify where our relaxations are loose. More broadly, we hope this work motivates the development of more low-compute and low-memory footprint solvers for real-time control.

\section*{Acknowledgments}
We thank Roy Xing, Gabriel Bravo, Moises Mata, and Charles Chen for help with hardware experiments. This work was supported by the National Science Foundation (under Awards 2144634, 2231350, and 2411369) and by the Center of AI Technology (CAIT) in collaboration with Amazon. Any opinions, findings, conclusions, or recommendations expressed in this material are those of the authors and do not necessarily reflect those of the funding organizations.

\bibliographystyle{plainnat}
\bibliography{inc/references}

\end{document}